\documentclass[conference]{IEEEtran}
\IEEEoverridecommandlockouts
\usepackage{cite}
\usepackage{amsmath,amssymb,amsfonts}
\usepackage{algorithmic}
\usepackage{graphicx}
\usepackage{textcomp}
\usepackage{xcolor}
\usepackage{hyperref}
\def\BibTeX{{\rm B\kern-.05em{\sc i\kern-.025em b}\kern-.08em
    T\kern-.1667em\lower.7ex\hbox{E}\kern-.125emX}}
\usepackage{stfloats}
\usepackage{comment}

\def\BibTeX{{\rm B\kern-.05em{\sc i\kern-.025em b}\kern-.08em
    T\kern-.1667em\lower.7ex\hbox{E}\kern-.125emX}}
\usepackage{tikz}
\newcommand*\emptycirc[1][0.7ex]{\tikz\draw (0,0) circle (#1);} 
\newcommand*\halfcirc[1][0.7ex]{%
  \begin{tikzpicture}
  \draw[fill] (0,0)-- (90:#1) arc (90:270:#1) -- cycle ;
  \draw (0,0) circle (#1);
  \end{tikzpicture}}
\newcommand*\fullcirc[1][0.7ex]{\tikz\fill (0,0) circle (#1);} 
\usepackage{booktabs}
    
\begin{document}
\title{A Graphical Modeling Language for Artificial Intelligence Applications in Automation Systems}
\author{
\IEEEauthorblockN{Marvin Schieseck, Philip Topalis, Alexander Fay}
\IEEEauthorblockA{
    \textit{Institute of Automation Technology} \\
    \textit{Helmut Schmidt University}\\
    Hamburg, Germany \\
    firstName.lastName@hsu-hh.de
    }
}
\maketitle

\begin{abstract} 
Artificial Intelligence (AI) applications in automation systems are usually distributed systems whose development and integration involve several experts. Each expert uses its own domain-specific modeling language and tools to model the system elements. An interdisciplinary graphical modeling language that enables the modeling of an AI application as an overall system comprehensible to all disciplines does not yet exist. As a result, there is often a lack of interdisciplinary system understanding, leading to increased development, integration, and maintenance efforts. This paper therefore presents a graphical modeling language that enables consistent and understandable modeling of AI applications in automation systems at system level. This makes it possible to subdivide individual subareas into domain specific subsystems and thus reduce the existing efforts.
\end{abstract}
 
\begin{IEEEkeywords}
AI System Modeling, Graphical Modeling Language, Domain Specific Modeling Language, Cyber Physical Systems
\end{IEEEkeywords}
\section{Introduction} \label{02_introduction}
The digitization and automation of technical plants, manufacturing processes, and products continue to increase and receive new impulses from modern information technologies. Companies are forced to adapt to new technologies such as Artificial Intelligence (AI) in order to remain competitive in the international market. \cite{OECD-2019-a}

For this reason, the integration of AI applications into automation systems has been increasingly pursued in recent years\cite{OECD-2019-b}. Numerous use cases have already been identified in which the integration of AI has increased the efficiency of previous solutions or made them possible in the first place \cite{Peres-2020}. Examples of this can be found in areas such as maintenance, quality control, or planning \cite{IEC-2018}.

Despite these promising use cases, AI is still rarely used in industry overall \cite{oecd-2019-d}. AI applications are often developed within research projects and to date, they can only be economically integrated to a limited extent in practice \cite{oecd-2019-c}. This is partly due to the fact that the software development of AI systems faces additional challenges compared to conventional software development, which hinder the economical further development and integration of experimental AI solutions\cite{Sculley-2015, hoffmann-2021}.

Sculley \textit{et al.}\cite{Sculley-2015} point out that most of these challenges need to be solved at the system level and not the code level. Furthermore, the authors highlight a lack of methods for abstract system modeling of AI applications. Kaymakci \textit{et al.} \cite{Kaymakci-2021}, Schröer \textit{et al.} \cite{Schroer-2021} and Saltz \cite{Saltz-2021} describe that there is no interdisciplinary communication framework for manufacturing companies that enables a common understanding of AI applications and defines basic functions and components. This research gap is summarized in \cite{IEC-2018} as well as \cite{acatech-req-2021}. 

Accordingly, there is a lack of interdisciplinary and easily understandable models for the development of AI applications. These models need to simultaneously capture the relationships between individual system components, software components, and system processes while supporting collaboration between different experts.

One way to address this gap is to provide a graphical representation of the system. Such a representation allows modeling the core of a problem in a formal and understandable way for different stakeholders\cite{Haberfellner-2019, VDI-3681}. In the field of automation technology and software engineering, graphical representations of systems have been frequently used for similar purposes for decades\cite{DSML-2013, Schnieder-1999, Stahl-2006}. A graphical modeling language (GML) enables the definition of symbols, rules, and semantics of such a representation, thus enabling understandable modeling for all experts involved\cite{VDI-3681}.

GMLs are divided into general-purpose modeling languages (GPMLs) and domain-specific modeling languages (DSMLs). GPMLs such as the Unified Modeling Language (UML) are extensive and flexible, allowing experts to model various problems from different technical domains based on general concepts such as classes, attributes, or inheritance. At the same time, this flexibility reduces the productivity of the modeling process and the understandability of the graphical representation because it requires reconstructing domain-level technical concepts from scratch based on the generic concepts. Instead, DSMLs are less flexible but provide domain concepts that correspond to the technical language and technical concepts used in the targeted domain. Hence, they offer the possibility to increase productivity and understandability, since the experts do not have to reconstruct the technical terms themselves from scratch. \cite{Frank-2010, Roychaudhuri-2019}

A suitable GML for the interdisciplinary modeling of AI applications in automation systems using domain-specific concepts does not yet exist. To close this gap, a new GML is presented. The main contributions of the paper are:
\begin{itemize}
    \item An evaluation of existing GMLs and their shortcomings for modeling AI applications in automation systems in Section \ref{03_related_works}.
    \item A presentation of the required system components, functions, and relations to model AI applications in automation systems in Section \ref{04_method}.
    \item A definition of symbols for these system elements and rules to relate them in Section \ref{04_method}.
    \item A validation of the GML based on a real-world example in Section \ref{05_validation}.
\end{itemize}

\section{Requirements and Related Works} \label{03_related_works}
Software development of AI applications faces additional challenges compared to traditional software development. These include unclear system boundaries, feedback loops, undeclared data dependencies, configuration issues, changes in the outside world, and system-level antipatterns, just to name a few. The code for the data-based AI models often consists of only a few lines of program code and represents only a small part compared to the total software code. The majority of an AI application consists of configuration, automation, testing, resource management, process and metadata management, deployment infrastructure, and data acquisition, data storage, data transmission, and data verification. \cite{Sculley-2015}

Another challenge is that any changes that cause future data to differ from historical data often will have a negative impact on the entire AI application. As a result, replacing, updating or calibrating hardware components such as sensors or actuators can have a negative impact on the overall AI application, even if it improves the overall data quality. \cite{Sculley-2015}

In addition, most AI applications in automation systems are decentralized systems, with various different software and hardware components distributed throughout the whole automation system\cite{Filho-2022}.

\subsection{Requirements for the Graphical Modeling Language}
Based on the presented challenges for the development of AI applications in automation systems, the requirements (R) for the new GML are derived in the following:
\subsubsection*{\textbf{R1: Interdisciplinary easy to understand}}
The GML has to be domain-independent and interdisciplinary easy to understand for all disciplines involved, in order to keep the effort and the hurdle for modeling low and to create a uniform communication basis\cite{acatech-req-2021, IEC-2018}.
\subsubsection*{\textbf{R2: Reduction of symbols}}
The GML has to represent information and components of the AI application, the automation system and the technical process necessary for the development and operation of the application in a clear and structured manner \cite{acatech-req-2021}. A formal reduction to a defined set of symbols to represent system components is required, as well as rules for their permissible connection \cite{Winzer-2016}. The GML has to provide a symbolically and semantically understandable representation that takes into account components, functions, relations\cite{VDI-3681, Haberfellner-2019}.
\subsubsection*{\textbf{R3: Represent interdependencies}}
The GML has to capture and represent technical and informational interdependencies between system components, AI components, and the technical process \cite{acatech-req-2021, Sculley-2015}.
\subsubsection*{\textbf{R4: Consideration of different AI software architectures}}
The GML has to model and represent different AI software architectures in order to consider different possible AI solutions \cite{Sculley-2015, Filho-2022}. 
\subsubsection*{\textbf{R5: Usable in parallel with process models for AI development}}
The GML must be able to be used in parallel with and integrated into common AI process models to minimize the additional effort required to develop AI applications \cite{Saltz-2021}.

\subsection{Process Models for AI Development}
For the development of complex technical systems, it is common to use process models. These models define a temporally logical sequence of actions and support project planning and the goal-oriented definition of the steps in a development process. \cite{Winzer-2016}

The \emph{Cross-Industry Standard Process for Data-Mining} (CRISP-DM) has become the de facto standard process model for various AI projects \cite{VDI-3714, Schroer-2021}. CRISP-DM is a domain-independent process model that divides the development process into six steps: \emph{Business Understanding, Data Understanding, Data Preparation, Modeling, Evaluation, and Deployment}\cite{Wirth-2000}. Iterations between individual steps are allowed. CRISP-DM is often used as a basis for the development of further domain specific process models \cite{Huber-2019, Studer-2021, Amershi-2019, Costa-2020, MartinezPlumed-2021}.

\subsection{Related Graphical Modeling Languages}
Kaymakci \textit{et al.} \cite{Kaymakci-2021} describe four elementary components of which an AI application is composed of and which can be used as a basis for the modeling: \emph{data source, data processing component, data-driven model, and AI service/agent}. For the graphical representation, rectangles and circles are used as symbols, which are labeled according to the components. These are connected by arrows to represent the flow of information and data. The data source records the data and sends it to other components. The data processing component converts its input data into output data of a defined format. The data-driven model also converts input data into output data, but only data-driven models are used for this, such as neural networks, which must be trained beforehand. The AI service/agent only uses the output of a data-driven model for the input data and converts it into output data or actions within the overall technical system.

The Standards Working Group for Measurement and Control Technology in the Chemical Industry (NAMUR) describes in \cite{NOA-2020} a concept for the development of Industrie 4.0 solutions, which also include AI applications. This concept is called \emph{NAMUR Open Architecture} (NOA). It is based on an extension of the classic automation pyramid from \cite{IEC-62264} and defines two additional separate areas: on the one hand a deterministic core process automation area and on the other hand a monitoring and optimization area. The graphical representation is done by rectangles, which are labeled according to the desired components. Each component is grouped into one of the two areas and arranged hierarchically based on the automation pyramid. An AI application is modeled by connecting the components from both areas through data interfaces using arrows. 

The Unified Modeling Language (UML) \cite{UML-Standard-v251} is a graphical modeling language, which is used in particular in the field of software engineering for modeling complex software systems. UML provides 14 graphical modeling types (diagrams) that can be used to create sketches and designs of a system. Each diagram has its own symbolism defined with its own rules and semantics.

The Systems Modeling Language (SysML) \cite{SysML-Standard-v16} is a graphical modeling language based on UML, which is used in particular in the field of systems engineering for modeling various complex technical systems. Like UML, SysML provides various graphical modeling types (diagrams), most of which are based on or derived from UML. Each diagram also has its own symbolism defined with its own rules and semantics.

The Association of German Engineers describes in VDI 3682\cite{VDI-3682} a graphical modeling language which is used to model technical processes throughout the life cycle of technical systems. The defined elements and symbols are product (circle), energy (rhombus), information (hexagon) and process operator (rectangle) and their flow (arrow with solid line) as well as technical resources (rounded rectangle) and assignments (double-headed arrow with dashed line). The process operator transforms products and energies. The flow of products, energies, and information between the process operators are represented by solid arrows. The process operator realizes the transformation with the help of the technical resource (rounded rectangle). Technical resources and process operators are assigned to each other by an assignment.

\subsection{Discussion of Related Works}
The established process models such as CRISP-DM support the planning of the entire development process and the development steps in the project, but not the actual design and modeling of the concrete AI application. None of the published process models adequately define how a distributed AI Application should be modeled and documented along each of the process steps. Nor does any of the process models refer to or recommend a graphical modeling language. \cite{Huber-2019, Studer-2021, Amershi-2019, Costa-2020, MartinezPlumed-2021, Wirth-2000}

Therefore, in the following, the previously presented modeling approaches are compared to the requirements in order to evaluate their suitability. This is summarized in Tab. \ref{tbl:evaluation table}.

The modeling according to Kaymakci \textit{et al.}\cite{Kaymakci-2021} consists of a few symbols and rules and is therefore rated as easy to understand (R1). The focus is on the representation of the AI application. A restriction on which basic components and functions have to be considered is only given for the AI application (R2). The system components or the technical process are not considered, except for the sensor. Therefore, the dependencies between the AI components themselves can be modeled, but not between the technical process or the plant components (R3). The modeling and consideration of different AI software architectures is not possible because the plant structure can not be taken into account (R4). The authors propose four process steps for modeling: \emph{Planning, Experimenting, Implementing} and \emph{Optimize}. A comparison of these steps with those in CRISP-DM shows that the approach can be integrated into established process models to a certain extent (R5).

Modeling according to NOA \cite{NOA-2020} consists of few symbols and rules and is therefore rated as easy to understand. Understandability is increased by the fact that the widely used automation pyramid is used as a basis (R1). A restriction on which basic components and functions must be taken into account is only made for individual plant components and a few AI components (R2). So in the NOA, individual AI and plant components are considered, but not the technical process. Therefore, the dependencies between AI components and plant components can be partially modeled, but not with reference to the technical process (R3). The representation of different AI software architectures is also limited, especially by separating them into the two domains. This predominantly supports the modeling of cloud architectures (R4). NOA does not further describe how the concept can be integrated into established process models (R5).

In principle, the graphical modeling with UML \cite{UML-Standard-v251} or SysML \cite{SysML-Standard-v16} using and combining different diagrams enables the modeling of mutual dependencies and of different software architectures (R3, R4). However, both are extensive modeling languages, which must first be learned by all participants before they can be understood and used \cite{Friedenthal-2012, Seidl-2015}. Each of the diagrams has its own symbols, rules, and semantics. 
In summary, therefore, these modeling languages are not rated as easy to understand for the development of AI applications (R1). The symbolism of the individual diagrams is defined, but not reduced to the important components for AI applications (R2). It is possible to integrate different diagrams into the individual steps of the established process models (R5).

The graphical modeling according to VDI 3682 \cite{VDI-3682} consists of few symbols and rules and is therefore rated as easy to understand (R1). The symbolism is reduced to plant components and the technical process, and thus does not consider software or AI components (R2). Interdependencies between the process and plant components can be captured, but not with respect to AI components. Software components and data flows as well as information flows can only be modeled to a limited extent and only partially represented graphically (R3). Therefore, the consideration of different AI software architectures is also not possible (R4). The modeling can be integrated for the Business Understanding step, but is hardly suitable for the following steps of CRISP-DM (R5).
\begin{table}[!h]
\renewcommand{\arraystretch}{1.25}
\caption{Evaluation of Related Graphical modeling languages}
\label{tbl:evaluation table}
\centering
    \begin{tabular}{l c c c c c}
    \toprule
       \textbf{Source} & \textbf{R1} & \textbf{R2} & \textbf{R3} & \textbf{R4} & \textbf{R5} \\
    \hline
        Kaymakci \textit{et al.} \cite{Kaymakci-2021}    & \fullcirc  & \halfcirc    & \halfcirc  & \emptycirc & \halfcirc \\
        NOA \cite{NOA-2020}         & \fullcirc  & \halfcirc    & \halfcirc  & \halfcirc   & \emptycirc \\
        UML \cite{UML-Standard-v251}, SysML \cite{SysML-Standard-v16}               & \emptycirc & \emptycirc   & \fullcirc  & \fullcirc   & \fullcirc \\
        VDI 3682 \cite{VDI-3682}         & \fullcirc  & \halfcirc    & \halfcirc  & \emptycirc  & \halfcirc\\
        \bottomrule
        \multicolumn{6}{c}{\fullcirc \: fulfilled \halfcirc \: partly fulfilled \emptycirc \: not fulfilled}
    \end{tabular}
\end{table}

Tab. \ref{tbl:evaluation table} shows that none of the existing approaches fulfills all requirements. Therefore, a new approach is needed, which is presented in the following.
\section{Graphical Modeling Language for Artificial Intelligence Applications in Automation Systems} \label{04_method}
According to Haberfellner \textit{et al.} \cite{Haberfellner-2019}, systems can be described generally using the following four elements: \textbf{\emph{System Components, System Functions, System Relations}} and \textbf{\emph{System Structures}}. For the new GML, these four elements are defined with the goal of fulfilling the listed requirements R1 to R5.

\subsection{System Components}
System components define the building blocks that the system comprises \cite{Haberfellner-2019}. According to the formulated requirements earlier, two types of system components are defined for the GML: Products and Technical Resources. Products are the things of the real world that are transformed during a technical process. Technical resources are the system components that build up the actual system and perform the system functions. For the GML, six possible categories of technical resources are defined:
\begin{itemize}
    \item Sensors: Devices that collect and digitize measured values from a technical process in the automation system.
    \item Actuators: Devices that intervene in a controlling manner in a process in the automation system. 
    \item Controllers: Devices which are responsible for the controlling of the behavior of the technical system. Signals are processed in short, generally deterministic, cycles. 
    \item Edge Devices: Micro controller or computers located locally in the automation system and thus close to the process.
    \item Local Computer Systems: Internal computer or data centers that are accessible via the corporate network without requiring an internet connection.
    \item Cloud Systems: External computer or data centers that can only be accessed via an Internet connection.
\end{itemize}
\subsection{System Functions}
System functions convert a given input into an output, thereby enabling the system to achieve its goals \cite{Haberfellner-2019}. These functions can be performed by technical resources. Seven categories of system functions are defined within the GML:
\begin{itemize}
        \item Automate: Read and process (sensor) values, and then control actuators to operate a technical process. This function describes that a technical process is automated by executing a program in the plant.
        \item Transform: Transforms real-world input products into output products in a technical process, for example a semifinished product into a workpiece through a machining process.
        \item Record: Captures and digitizes physical properties (pressure, temperature). The record function also includes the recording of non-physical properties, such as time information or economic information (prices, stocks).
        \item Store: Stores data according to a certain system, for a certain time at a certain place.
        \item Process: Processes and transforms input data or measured values into output data by a defined algorithm. An example is data pre-processing or post-processing. No data-driven algorithm may be used.
        \item Train: Adjusts the parameters of a data-driven algorithm based on historical data.
        \item Inference: Converts input data into output data using a data-driven algorithm. A training must have already adjusted the data-driven algorithm.
\end{itemize}
\subsection{System Relations}
System relations describe the connections and relationships between system elements \cite{Haberfellner-2019}. This allows interdependencies to be modeled, identified and represented. Three categories of system relations are defined within the GML:
\begin{itemize}
    \item Communication: A connection between at least two system components that describes the directed exchange of information between these. An example of this is the communication between a controller and a sensor via TCP/IP.
    \item Assignment: A structural connection between system functions and system components. It assigns one system function to one system component, for example a system function record could be assigned to a system component sensor. An exception is the transform function, which can be assigned to multiple system components.
    \item Product Flow: A connection that describes the directed flow of a product between transformation functions. 
\end{itemize}
\subsection{System Structure}
The system structure defines an ordering principle according to how the individual components of the system are ordered and hierarchized \cite{Haberfellner-2019}. In automation engineering, the automation pyramid is an accepted, widely used and easily understandable way to order and hierarchize an automation system \cite{IEC-62264}. Following this, a four-level system structure is defined for the GML:
\begin{itemize}
    \item Level\:1\:-\:Technical Process Level:
    Here, the transformation processes of the products are described. 
    \item Level\:2\:-\:Field Device Level: 
    Here, the sensors and actuators are described.  
    \item Level\:3\:-\:Control and Monitoring Level: 
    Here, the control systems and edge devices are described.
    \item Level\:4\:-\:Computer and Cloud Level: 
    Here, the local and external computer and cloud systems are described.
\end{itemize}
\subsection{Standardized Graphical Symbols}
Standardized symbolism of the defined system elements allows simplifying the development process of an AI application. In addition, the results of the development can be communicated and documented more precisely. \cite{Winzer-2016, VDI-3681}

\begin{figure}[!ht]
    \centering
    \includegraphics{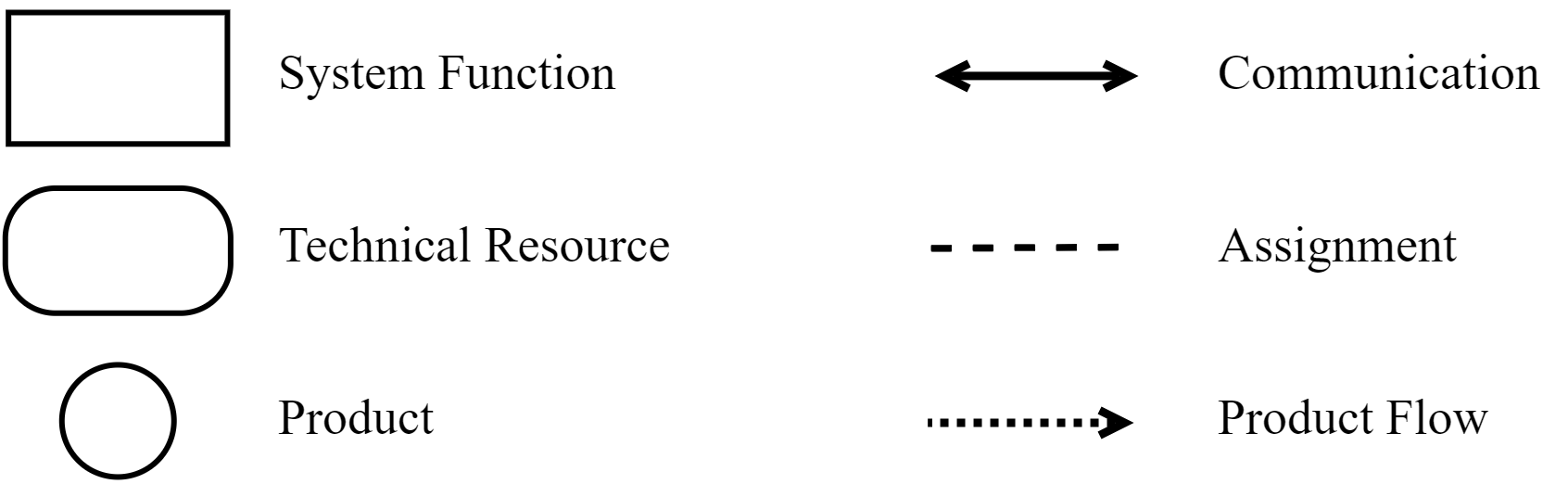} 
    \caption{Overview of standardized symbols for graphical modeling language.}
    \label{fig:bm_symbols}
\end{figure}

Therefore, a standardized symbolism is defined for the GML. Rectangles are used for the system functions. Rectangles with rounded corners are used for the technical resources and circles are used for the products. A communication is represented by a solid line, while an arrow defines the communication direction. Two arrows represent a bidirectional communication. An assignment is represented by a dashed line without an arrow. A product flow is represented by a dotted line, while an arrow defines the flow direction. An overview of those symbols is shown in Fig. \ref{fig:bm_symbols}.
\subsection{Methodology for using the Graphical Modeling Language}
In order for the GML to be integrated into the established process models like CRISP-DM, it has to support its individual steps. So, in the first step of CRISP-DM, the business and project goals are defined. An example of this could be the reduction of maintenance costs for a part of the automation system. At the same time, technical contexts are also considered and compared with the business goal in order to define technically relevant parameters. An example of this could be that a large part of the cost of maintaining a machine is due to the repair of a particular component. During this step, the graphical modeling language can be used to model the current state of the automation system and the technical process. This allows the problem space to be formalized, documented, and communicated among all stakeholders and disciplines. Subsequently, a solution space can be developed collectively by all stakeholders. In the further individual steps of the CRISP-DM, the graphical model of the current state can be successively extended and adapted to develop the AI application. The following methodology is recommended for using the graphical modeling language:
First, the current state of the automation system should be modeled along the mentioned levels, from bottom to top. Therefore, modeling should start at the process level, extend to the field device and control level, and end at the computer and cloud system level if present. At each level, the system components, relations, and functions that influence the technically relevant parameters should be modeled. After modeling the current state, possible solution concepts for the AI application should then be modeled. This is done by adding additional system components, functions, and relations or by modifying the existing ones. Depending on the combination and distribution of functions, components, and relationships, different AI software architectures can be designed and discussed among all stakeholders and disciplines.

\section{Validation} \label{05_validation}
The validation is done by modeling an AI application for a real-world use case using the GML. The business goal of the presented use case is to reduce the maintenance costs of a stamping machine. The machine consists of a support frame, a fixed lower die and a movable upper die. The upper die is moved by an electric motor, which is mounted on the support frame. The driving force is transmitted via the drive belt and the position of the upper die is measured via two positional sensors. This machine performs a stamping process in which a metal plate is processed. The technical goal is to optimize its maintenance strategy by predicting the wear condition of the drive belt. Currently, the drive belt is maintained in fixed intervals.
Some interviews with process operators revealed that a worn-out drive belt causes an oscillation of the upper stamp die when it gets positioned. Based on that assumption, the measured positioning values recorded during the stamping process can be used to infer the belt wear condition. For the automated analysis of this position data, an AI application should be developed.

\subsubsection*{Modeling of the Process Level}
At the beginning of the manufacturing process, a metal plate is inserted into the lower die. The upper die is then lowered from its initial position until the machine is closed. While closing, the upper die punches out a part from the raw metal plate. After that, the tool opens and the upper die drives back to its initial position. Then a worker removes the finished part and the rest material. This process layer is shown in Fig. \ref{fig:bm_maintenance} in layer 1.
\subsubsection*{Modeling of the Field Device Level}
The two position sensors measure the position of the upper die during the closing and opening process. Therefore, one recording function is assigned to every position sensor. In Fig. \ref{fig:bm_maintenance}, this is shown in layer 2.
\subsubsection*{Modeling of the Control and Monitoring Level}
The position sensors transmit the position data to the motor controller. An automation function is assigned to the motor controller, which performs the direct control of the motor and thus the positioning of the upper die. The machine controller coordinates the manufacturing process, for example by setting up the machine parameters and starting or stopping the manufacturing process. Therefore, the machine controller is also assigned to an automate function. The machine controller communicates with the motor controller and vice versa. Furthermore, the machine controller records some time stamps. Therefore, it also has a recording function assigned to. In Fig. \ref{fig:bm_maintenance}, those components are added to layer 3. By modeling these components functions and relationships, the current state of the automation system is represented.
\subsubsection*{Modeling of the Solution Concept}
\begin{figure}[ht]
    \centering
    \includegraphics{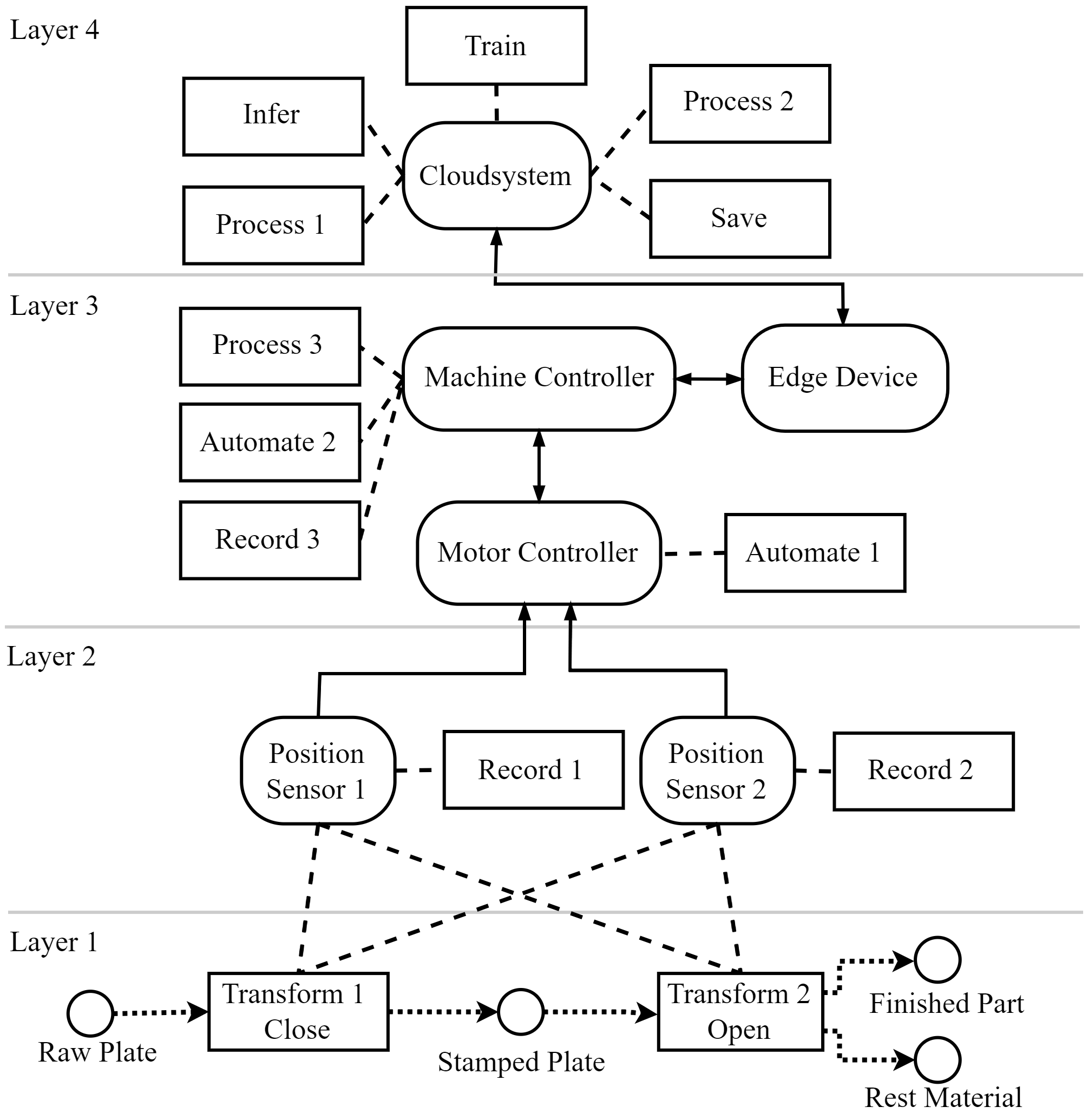} 
    \caption{Graphical representation of the solution concept using the GML.}
    \label{fig:bm_maintenance}
\end{figure}

It is sufficient for the business use case, if the wear condition of the belt can be queried on demand every few minutes. This is due to the fact that the wear condition of the belt slowly decreases, at least over several hours. Therefore, it is not required that the AI application is executed in real time with respect to the manufacturing process. As a result, a cloud architecture is chosen for the solution concept. The data is stored and pre-processed in the cloud. The training of the model and the inference are also performed in the cloud. The result of the inference is post-processed so that it can be queried by the machine controller. In summary, store, inference, train, and process functions are assigned to the cloud. The machine controller and the cloud communicate via an edge device and thus exchange measured positional data and wear condition inference results. The inference results are shown by the machine controller on a display to the operator. For this purpose, a process function is assigned to the machine controller. The modeled solution concept for the AI application is shown in Fig. \ref{fig:bm_maintenance}.
\section{Discussion} \label{06_discussion}
The GML defines a symbolism and rules that can be easily understood across disciplines. This was shown through the modeling of the use case. The presented methodology, which recommended modeling bottom up along the levels, additionally enhanced the interdisciplinary understanding (R1). 

The defined symbolism allows representing the AI components, the automation system structure and the process. Nevertheless, the modeling is limited to the relevant elements at the same time. (R2)

By linking technical resources with communications, or assigning functions to technical resources or assigning technical resources to manufacturing process steps, the interdependencies of the automation system and the AI application are represented. For example, if a position sensor is replaced in the use case, it is immediately apparent that the AI application is affected by changes in the recording function. (R3)

The GML enables the modeling of different AI software architectures because the AI functions can be assigned to any technical resource depending on the use case requirements. Obviously, an alternative AI architecture could have been modeled instead, such as an edge or hybrid architecture. (R4)

The GML methodology supports CRISP-DM in the initial steps of problem formulation and in the later steps of solution finding, comparison and documentation. The easy graphical representation helps to build a cross-domain understanding of the system and enables the distribution of tasks to individual stakeholders along the CRISP-DM. (R5)

The interfaces between all functions as well as the distribution of the software to the hardware are recognizable for all stakeholders involved. In addition, the GML eliminates the need for domain experts to reconstruct domain-level technical concepts from scratch using generic concepts, which would be required when using UML or SysML. Based on that, the GML enables a new way of standardized documenting AI applications in automation systems, which improves its subsequent integration and long-term maintenance in the lifecycle of the automation system. 

In its current form, the GML cannot model or represent time dependencies between functions, for example, that data pre-processing needs to be done before inference. Furthermore, there is currently no possibility of assigning attributes to the individual system elements. This would allow important system attributes to be documented and displayed, such as used communication speed, used communication protocols or used data-driven algorithms.
\section{Conclusion} \label{07_conclusion}
In this paper, a graphical modeling language (GML) for Artificial Intelligence (AI) applications in automation systems was presented. It allows modeling AI applications for automation systems in an interdisciplinary and easily understandable way. For the GML, a symbolism as well as rules for the combination of the symbols and their semantics were defined. It defines functions and components, which are linked to each other via relationships. That allows a graphical representation of automation systems and  AI applications with its elements, even for different AI system architectures like cloud or edge architectures. In addition, the interfaces between the elements as well as the distribution of the software to the hardware becomes directly visible to all participants and experts involved. Therefore, the graphical representation provides an interdisciplinary communication framework for cross-domain collaboration, which promotes communication while it documents the solution understandably at the same time. The shown GML can be used in conjunction and parallel with common process models such as CRISP-DM. This keeps the additional effort of using the GML to a minimum.

Overall, modeling based on this new GML offers the potential to facilitate the development, integration, operation, and documentation as well as the maintenance of AI applications in automation systems.
\section{Future works} \label{08_future_work}
For the future, it is planned to extend the validation. First, by modeling more real-world use cases and second, by conducting surveys among domain experts to validate the fulfillment of the presented requirements.

In the next step, the GML will be extended by an information and meta-model using the Meta Object Facility (MOF) \cite{MOF-Standard-v251}. This would allow to automate consistency or rule based checks of the modeled system. Moreover, it would allow functions such as data preprocessing to be systematically encapsulated in the form of software components and thus made reusable. Finally, it could enable systematic categorization, comparison, and search of different use cases and solutions that have already been modeled. The information model, along with the meta-model, are provided at \cite{gml-aiaas}.

\section*{Acknowledgements}
This research is funded by dtec.bw – Digitalization and Technology Research Center of the Bundeswehr which we gratefully acknowledge. dtec.bw is funded by the European Union – NextGenerationEU.

\bibliographystyle{IEEEtran}
\bibliography{references.bib}

\begin{thebibliography}{10}
\providecommand{\url}[1]{#1}
\csname url@samestyle\endcsname
\providecommand{\newblock}{\relax}
\providecommand{\bibinfo}[2]{#2}
\providecommand{\BIBentrySTDinterwordspacing}{\spaceskip=0pt\relax}
\providecommand{\BIBentryALTinterwordstretchfactor}{4}
\providecommand{\BIBentryALTinterwordspacing}{\spaceskip=\fontdimen2\font plus
\BIBentryALTinterwordstretchfactor\fontdimen3\font minus
  \fontdimen4\font\relax}
\providecommand{\BIBforeignlanguage}[2]{{%
\expandafter\ifx\csname l@#1\endcsname\relax
\typeout{** WARNING: IEEEtran.bst: No hyphenation pattern has been}%
\typeout{** loaded for the language `#1'. Using the pattern for}%
\typeout{** the default language instead.}%
\else
\language=\csname l@#1\endcsname
\fi
#2}}
\providecommand{\BIBdecl}{\relax}
\BIBdecl

\bibitem{OECD-2019-a}
\BIBentryALTinterwordspacing
``\BIBforeignlanguage{en}{{Economic Outlook, Volume 2019 Issue 1}},''
  {Organisation for Economic Co-operation and Development (OECD)}, Tech. Rep.,
  May 2019. [Online]. Available: \url{https://doi.org/10.1787/b2e897b0-en}
\BIBentrySTDinterwordspacing

\bibitem{OECD-2019-b}
\BIBentryALTinterwordspacing
``\BIBforeignlanguage{en}{{Artificial Intelligence in Society}},''
  {Organisation for Economic Co-operation and Development (OECD)}, Tech. Rep.,
  Jan. 2019. [Online]. Available: \url{https://doi.org/10.1787/eedfee77-en}
\BIBentrySTDinterwordspacing

\bibitem{Peres-2020}
R.~S. Peres, X.~Jia, J.~Lee, K.~Sun, A.~W. Colombo, and J.~Barata,
  ``{Industrial Artificial Intelligence in Industry 4.0-Systematic Review,
  Challenges, and Outlook},'' \emph{IEEE Access}, vol.~8, 2020.

\bibitem{IEC-2018}
``\BIBforeignlanguage{en}{{Artificial Intelligence Across Industries}},''
  {International Electrotechnical Commission (IEC)}, Tech. Rep., Oct. 2018.

\bibitem{oecd-2019-d}
\BIBentryALTinterwordspacing
M.~Bianchini and V.~Michalkova, ``Data analytics in smes,'' {{Organisation for
  Economic Co-operation and Development (OECD)}}, Tech. Rep.~15, 2019.
  [Online]. Available:
  \url{https://doi.org/https://doi.org/10.1787/1de6c6a7-en}
\BIBentrySTDinterwordspacing

\bibitem{oecd-2019-c}
\BIBentryALTinterwordspacing
``{The Digitalisation of Science, Technology and Innovation},'' {Organisation
  for Economic Co-operation and Development (OECD)}, Tech. Rep., Feb. 2020.
  [Online]. Available: \url{https://doi.org/10.1787/b9e4a2c0-en}
\BIBentrySTDinterwordspacing

\bibitem{Sculley-2015}
D.~Sculley, G.~Holt, D.~Golovin, E.~Davydov, T.~Phillips, D.~Ebner,
  V.~Chaudhary, M.~Young, J.-F. Crespo, and D.~Dennison, ``{Hidden Technical
  Debt in Machine Learning Systems},'' in \emph{Proceedings of the 28th
  International Conference on Neural Information Processing Systems - Volume
  2}, ser. NIPS'15.\hskip 1em plus 0.5em minus 0.4em\relax MIT Press, 2015, p.
  2503–2511.

\bibitem{hoffmann-2021}
M.~W. Hoffmann, R.~Drath, and C.~Ganz, ``Proposal for requirements on
  industrial ai solutions,'' in \emph{{Machine Learning for Cyber Physical
  Systems}}, J.~Beyerer, A.~Maier, and O.~Niggemann, Eds.\hskip 1em plus 0.5em
  minus 0.4em\relax Springer Berlin Heidelberg, 2021, pp. 63--72.

\bibitem{Kaymakci-2021}
C.~Kaymakci, S.~Wenninger, and A.~Sauer, ``{A Holistic Framework for AI Systems
  in Industrial Applications},'' in \emph{Innovation Through Information
  Systems}.\hskip 1em plus 0.5em minus 0.4em\relax Springer International
  Publishing, 2021, pp. 78--93.

\bibitem{Schroer-2021}
C.~Schröer, F.~Kruse, and J.~M. Gómez, ``{A Systematic Literature Review on
  Applying CRISP-DM Process Model},'' \emph{Procedia Computer Science}, vol.
  181, pp. 526--534, 2021.

\bibitem{Saltz-2021}
J.~S. Saltz, ``{CRISP-DM for Data Science: Strengths, Weaknesses and Potential
  Next Steps},'' in \emph{2021 IEEE International Conference on Big Data (Big
  Data)}.\hskip 1em plus 0.5em minus 0.4em\relax IEEE, Dez 2021, pp.
  2337--2344.

\bibitem{acatech-req-2021}
``\BIBforeignlanguage{de}{{Modellierung- und Simulationsbedarfe der
  intelligenten Fabrik}},'' {Forschungsbeirat der Plattform Industrie 4.0 /
  acatech - Deutsche Akademie der Technikwissenschaften}, Tech. Rep., Dez.
  2021.

\bibitem{Haberfellner-2019}
R.~Haberfellner, O.~de~Weck, E.~Fricke, and S.~V{\"o}ssner, \emph{{Systems
  Engineering}}.\hskip 1em plus 0.5em minus 0.4em\relax {Springer International
  Publishing}, 2019.

\bibitem{VDI-3681}
\emph{{Classification and Evaluation of Description Methods in Automation and
  Control Technology}}, {Verein Deutscher Ingenieure e.V. (VDI)} Std. 3681,
  Oct. 2005.

\bibitem{DSML-2013}
U.~Frank, \emph{Domain-Specific Modeling Languages: Requirements Analysis and
  Design Guidelines}, I.~Reinhartz-Berger, A.~Sturm, T.~Clark, S.~Cohen, and
  J.~Bettin, Eds.\hskip 1em plus 0.5em minus 0.4em\relax Berlin, Heidelberg:
  Springer Berlin Heidelberg, 2013.

\bibitem{Schnieder-1999}
E.~Schnieder, \emph{{Methoden der Automatisierung}}.\hskip 1em plus 0.5em minus
  0.4em\relax Wiesbaden: {Vieweg+Teubner Verlag}, 1999.

\bibitem{Stahl-2006}
T.~Stahl, M.~Völter, J.~Bettin, A.~Haase, and S.~Helsen, \emph{{Model-Driven
  Software Development: Technology, Engineering, Management}}.\hskip 1em plus
  0.5em minus 0.4em\relax Wiley, Jan. 2006.

\bibitem{Frank-2010}
U.~Frank, ``Outline of a {{Method}} for {{Designing Domain-Specific Modelling
  Languages}},'' Tech. Rep., 2010.

\bibitem{Roychaudhuri-2019}
S.~Roy~Chaudhuri, S.~Natarajan, A.~Banerjee, and V.~Choppella, ``Methodology to
  develop domain specific modeling languages,'' in \emph{Proceedings of the
  17th {{ACM SIGPLAN International Workshop}} on {{Domain-Specific
  Modeling}}}.\hskip 1em plus 0.5em minus 0.4em\relax {Athens Greece}: {ACM},
  Oct. 2019, pp. 1--10.

\bibitem{Filho-2022}
C.~P. Filho, E.~Marques, V.~Chang, L.~dos Santos, F.~Bernardini, P.~F. Pires,
  L.~Ochi, and F.~C. Delicato, ``{A Systematic Literature Review on Distributed
  Machine Learning in Edge Computing},'' \emph{Sensors}, vol.~22, 2022.

\bibitem{Winzer-2016}
P.~Winzer, \emph{{Generic Systems Engineering}}.\hskip 1em plus 0.5em minus
  0.4em\relax {Springer Berlin Heidelberg}, 2016.

\bibitem{VDI-3714}
\emph{{Implementation and operation of Big Data application in the
  manufacturing industry - Implementation of Big Data projects}}, {Verein
  Deutscher Ingenieure e.V. (VDI)} Std. 3714-1, Sep. 2022.

\bibitem{Wirth-2000}
R.~Wirth and J.~Hipp, ``{CRISP-DM: Towards a Standard Process Model for Data
  Mining},'' \emph{Proceedings of the 4th International Conference on the
  Practical Applications of Knowledge Discovery and Data Mining}, Jan. 2000.

\bibitem{Huber-2019}
S.~Huber, H.~Wiemer, D.~Schneider, and S.~Ihlenfeldt, ``{DMME: Data Mining
  Methodology for Engineering Applications – a Holistic Extension to the
  CRISP-DM Model},'' \emph{Procedia CIRP}, vol.~79, pp. 403--408, Jan. 2019.

\bibitem{Studer-2021}
S.~Studer, B.~Bui, C.~Drescher, A.~Hanuschkin, L.~Winkler, S.~Peters, and K.-R.
  Müller, ``{Towards CRISP-ML(Q): A Machine Learning Process Model with
  Quality Assurance Methodology},'' \emph{Machine Learning and Knowledge
  Extraction}, vol.~3, pp. 392--413, Apr. 2021.

\bibitem{Amershi-2019}
S.~Amershi, A.~Begel, C.~Bird, R.~DeLine, H.~Gall, E.~Kamar, N.~Nagappan,
  B.~Nushi, and T.~Zimmermann, ``{Software Engineering for Machine Learning: A
  Case Study},'' in \emph{2019 IEEE/ACM 41st International Conference on
  Software Engineering: Software Engineering in Practice (ICSE-SEIP)}, 2019,
  pp. 291--300.

\bibitem{Costa-2020}
C.~J. Costa and J.~T. Aparicio, ``{POST-DS: A Methodology to Boost Data
  Science},'' in \emph{2020 15th Iberian Conference on Information Systems and
  Technologies (CISTI)}.\hskip 1em plus 0.5em minus 0.4em\relax IEEE, 2020.

\bibitem{MartinezPlumed-2021}
F.~Martinez-Plumed, L.~Contreras-Ochando, C.~Ferri, J.~Hernandez-Orallo,
  M.~Kull, N.~Lachiche, M.~J. Ramirez-Quintana, and P.~Flach, ``{CRISP-DM
  Twenty Years Later: From Data Mining Processes to Data Science
  Trajectories},'' \emph{IEEE Transactions on Knowledge and Data Engineering},
  vol.~33, no.~8, pp. 3048--3061, 2021.

\bibitem{NOA-2020}
\emph{{NAMUR Open Architecture - NOA Concept}}, {Normenarbeitsgemeinschaft für
  Meß- und Regeltechnik in der chemischen Industrie (NAMUR)} Std. NE 175, Jul.
  2020.

\bibitem{IEC-62264}
\emph{{Enterprise-control System Integration}}, {International Electrotechnical
  Commission (IEC)} Std. 62\,264, May 2013.

\bibitem{UML-Standard-v251}
\BIBentryALTinterwordspacing
\emph{{Unified Modeling Language (UML) Version 2.5.1}}, {Object Management
  Group ({OMG})} Std., Dec. 2017. [Online]. Available:
  \url{https://www.omg.org/spec/UML/2.5.1}
\BIBentrySTDinterwordspacing

\bibitem{SysML-Standard-v16}
\BIBentryALTinterwordspacing
\emph{{Systems Modeling Language (SysML) Version 1.6}}, {Object Management
  Group (OMG)} Std., Dec. 2019. [Online]. Available:
  \url{https://www.omg.org/spec/SysML/1.6}
\BIBentrySTDinterwordspacing

\bibitem{VDI-3682}
\emph{{Formalised Process Descriptions}}, {Verein Deutscher Ingenieure e.V.
  (VDI)} Std. 3682, May 2005.

\bibitem{Friedenthal-2012}
S.~Friedenthal, A.~Moore, and R.~Steiner, \emph{{A Practical Guide to SysML:
  Systems Modeling Language}}.\hskip 1em plus 0.5em minus 0.4em\relax Morgan
  Kaufmann Publishers Inc., 2008.

\bibitem{Seidl-2015}
M.~Seidl, M.~Scholz, C.~Huemer, and G.~Kappel, \emph{{UML @ Classroom}}.\hskip
  1em plus 0.5em minus 0.4em\relax {Springer International Publishing}, 2015.

\bibitem{MOF-Standard-v251}
\BIBentryALTinterwordspacing
\emph{{OMG Meta Object Facility (MOF) Core Specification Version 2.5.1}},
  {Object Management Group ({OMG})} Std., Oct. 2019. [Online]. Available:
  \url{https://www.omg.org/spec/MOF/2.5.1/PDF}
\BIBentrySTDinterwordspacing

\bibitem{gml-aiaas}
\BIBentryALTinterwordspacing
M.~Schieseck. {A Graphical Modeling Language for Artificial Intelligence
  Applications in Automation Systems}. {[Accessed: 05-Jun-2023]}. [Online].
  Available: \url{{https://github.com/schiesem/GML-AIAAS}}
\BIBentrySTDinterwordspacing

\end{thebibliography}
\end{document}